# A Comparison of Feature-Based and Neural Scansion of Poetry

**Manex Agirrezabal**[1] and **Iñaki Alegria**[1] and **Mans Hulden**[2]

IXA NLP Group[1]
Department of Computer Science
Univ. of the Basque Country (UPV/EHU)
`manex.aguirrezabal@ehu.eus`
`i.alegria@ehu.eus`

Department of Linguistics[2]
University of Colorado
`mans.hulden@colorado.edu`

## Abstract

Automatic analysis of poetic rhythm is a challenging task that involves linguistics, literature, and computer science. When the language to be analyzed is known, rule-based systems or data-driven methods can be used. In this paper, we analyze poetic rhythm in English and Spanish. We show that the representations of data learned from character-based neural models are more informative than the ones from hand-crafted features, and that a Bi-LSTM+CRF-model produces state-of-the art accuracy on scansion of poetry in two languages. Results also show that the information about whole word structure, and not just independent syllables, is highly informative for performing scansion.

## 1 Introduction

*I don't like to brag and I don't like to boast*[1]
*Questi non ciberà terra né peltro,*[2]
Мой дядя самых честных правил,[3]

The above are examples of metered poetry in English, Italian and Russian. If the English example is read out loud, it is probably rendered in a continuous deh-deh-**dum** pattern. In the second example, the line consists of eleven beats where some syllables (in fixed positions) are more prominent than others.[4] The Russian example is part of a poem written completely in iambic meter (using a recurring deh-**dum** sound pattern).[5] A person able to read texts in Russian would most likely produce this recurring pattern when reciting the poem. A far more interesting question is whether this rhythmic structure of the poem can be discovered without possessing complete understanding of the language. Or, whether we could even analyze it without any knowledge of the language in question. These are a difficult challenge for NLP that involve knowledge about linguistics, literature and computer science.

To understand the underlying prosodic structure of a poem independently of the language, a necessary core piece of knowledge concerns the typological relationship between different poetic traditions. This work represents the first steps towards an understanding of how to incorporate such knowledge into practical systems. To this end, we scan[6] the rhythm of poems using data-driven techniques with two languages.[7] In our previous work we tested basic techniques on English poetry (Agirrezabal et al., 2016a); in this research we improve the results using deep learning and extend the experiments to include Spanish poetry. The analysis of the results and adopting our models to perform fully unsupervised and language independent poetry analysis is our current challenge.

## 2 Scansion

Performing scansion of a line of poetry involves marking the rhythmic structure of that line, along with feet (groups of syllables) and rhyme patterns across lines (Corn, 1997; Fabb, 1997; Steele, 1999). In this work, however, we address only the task of inferring the stress sequence for each verse (a sequence of words or syllables).

### 2.1 English

Poems in English contain repeating patterns of syllable stress groupings, better known as feet, and according to the type of foot used, i.e. the number of syllables in each, several meters can be employed. The most common ones are iambic feet (bal-**loon**), trochaic (**jun**-gle), dactylic (**ac**-ci-dent) or anapestic (com-pre-**hend**).

---

[1] Dr. Seuss' *Scrambled Eggs Super!*
[2] Dante Alighieri's *The Divine Comedy* (Canto I, Inferno).
[3] Alexander Pushkin's *Eugene Onegin*.
[4] As a rule of thumb, the 10th beat is always stressed.
[5] A complete reading of each poem should convince the reader that this pattern is present throughout.

[6] The common term for annotating poetry with stress levels.
[7] The repository with the data and techniques: https://github.com/manexagirrezabal/herascansion/

The length of a metrical line is expressed by the number of feet found in regular lines. Thus a dimeter has two feet, a trimeter three, a tetrameter four, and so on (pentameter, hexameter, heptameter,...). The most common meter in English is iambic pentameter, e.g.

> *O change | thy **thought**, | that **I**  
> may **change** | my **mind**,*

Although poems show an overall regularity throughout lines, poets tend to vary some parts of verse slightly, with various artistic motives for doing so, as in

> *Grant, if thou wilt, thou art beloved of many*

This differs from the previous example by its prominent **dum**-deh-deh-**dum** pattern early on—**grant**, if thou **wilt**—known in the literature as a 'trochaic variation'. Another variation is that, since the poem is iambic overall, the final syllable in the line should be stressed, but it instead ends with an unstressed **ny**-syllable. Appending an unstressed syllable at the end of an iambic line is a common departure of a set form in English poetry called *feminine ending*. An automated scansion system must be aware of, or learn, such common variants and be able to apply them consistently.

### 2.2 Spanish

In the Spanish poetic tradition, several metrical structures have been popular over time (Quilis, 1984; Tomás, 1995; Caparrós, 1999). In this work, because of corpus availability, we have only focused on a specific time period, the Golden Age. In this period the main meter of poetry was the hendecasyllable, in which each line of verse consists of eleven syllables. The stress sequence is quite regular and usually the 10th syllable is stressed. Other syllable positions are also stressed and the specifics of the pattern leads to a rich categorization of hendecasyllabic lines, which is outside our current scope of work.

One of the challenges in analyzing Spanish poetry is the use of syllable contractions, also known as synalephas, to force verses with more than eleven syllables into hendecasyllabic structures. Because of this, when scansion is performed, not all the syllables receive a stress value. As one of our intentions was to reproduce the experiments and methods of previous work, we have created a heuristic to assign a stress value to each syllable (by adding unstressed syllables and maintaining lexical stresses when possible).

### 2.3 Automated scansion

Automated scansion is a vibrant topic of research. Recent work often casts this as a prediction problem, where receiving a sequence of words in a poem as input we must predict the stress patterns for each of them. This prediction is often approached in one of two different ways; either following expert-designed **rules** that guide the marking, or learning from patterns in labeled data. Rule-based work include Logan (1988); Hartman (2005); Plamondon (2006); McAleese (2007); Gervas (2000); Navarro-Colorado (2015); Agirrezabal et al. (2016b). Currently, data-driven techniques are becoming more popular due to the availability of tagged data. Some works that employ data and get information from it are Hayward (1996); Greene et al. (2010); Hayes et al. (2012); Agirrezabal et al. (2016a); Estes and Hench (2016).

## 3 Corpora

As the gold standard material for training the English metrical tagger, we used a corpus of scanned poetry, For Better For Verse (4B4V), from the University of Virginia (Tucker, 2011).[8] The entire collection consists of 78 poems, approximately 1,100 lines in total. Sometimes several analyses are given as correct, as there is some natural ambiguity when performing scansion—about 10% of the lines are ambiguous with two or more plausible analyses given.

For the Spanish language portion we make use of a corpus of Spanish Golden-Age Sonnets (Navarro-Colorado et al., 2016) available on GitHub.[9] This is a collection of poems from the 16th and 17th centuries, which has been manually checked, contains approximately 135 sonnets and almost 2,000 lines. These poems were written by seven different well-known authors.

---

[8] http://prosody.lib.virginia.edu/  
[9] https://github.com/bncolorado/CorpusSonetosSigloDeOro

| |
|---|
| English |
| *The **jaws** that **bite**, the **claws** that **catch**!* <br> Eight segments, four strong beats |
| Spanish |
| *su **fá**brica en tus **ru**inas ade**lan**ta,* <br> Eleven segments, three strong beats |

## 4 Methods

We follow the intuitions outlined in Agirrezabal et al. (2016a) and we use the same set of linguistically motivated features. The feature templates include current and surrounding words, syllables, POS-tags and lexical stresses, among other simpler ones. This paper extends the work as more current methods—neural network models in particular—and a new language is explored.

The earlier feature-based systems require manual extraction of features where for each syllable in the dataset we extract a set of 10 basic feature templates extended by another set of 54 feature templates. Neural network based methods do not need this feature extraction phase.

We have extended the methods and frameworks presented in Agirrezabal et al. (2016a) to analyze verses in the two datasets. The algorithms include the Averaged Perceptron,[10] (Rosenblatt, 1958; Freund and Schapire, 1999), Hidden Markov Models (Rabiner, 1989; Halácsy et al., 2007), and Conditional Random Fields (CRFs)[11] (Lafferty et al., 2001; Okazaki, 2007). Beyond this, we also performed further experiments by employing Bidirectional LSTMs with a CRF layer (Lample et al., 2016).[12]

Initially, we performed preliminary experiments using an encoder-decoder model[13] (Bahdanau et al., 2014; Kann and Schütze, 2016) and also Recurrent Neural Network Language Models[14] (Mikolov et al., 2010), but these performed less well in our experiments.

The specific Bi-LSTM+CRF model from Lample et al. (2016) is an architecture that is suitable for our problem.[15] Words are modeled with a character-based RNN with LSTM, which produces two vectors. The forward vector will have a representation of the character sequence from the left to the right. The backward one will have the same in the reversed order. Our insight is that this character-based LSTM captures the phonological structure of the word from its graphemes/characters. These two vectors are concatenated together with the whole word's embedding (the embeddings could be pre-trained from larger corpora or trained jointly for the task). The vector of these three elements will represent each word in the sequence. Then, for each word, there will be a word-level LSTM, which will produce an output for each word, with its right and left context information. Finally, this output will go through a CRF layer to get the optimal output. For details, we refer the reader to Lample et al. (2016).

We performed several experiments. In some cases, the models were designed to learn a direct mapping from syllables to stresses (**S2S**[16]). In other cases, for each syllable we extracted its respective feature templates (10 or 64) and learned from that data (S2S with more features). With the neural model, the dataset consisted of sequences of words or syllables and the framework had to infer the output (the stress). If the input was a sequence of words, as some words can have more than one syllable, the output had to be a stress pattern, and not only a single stress value (**W2SP**[17]). We decided to use this learning mode to check if the inclusion of independently pre-trained word embeddings would improve our results.[18] When the input was a sequence of syllables separated by spaces, word structure information could be lost. In order to handle this, we included word boundary markers (**WB**) in some experiments.

## 5 Evaluation and Results

We performed a 10-fold cross-validation to evaluate our models, due to the small size of the tagged datasets.

In assessing each of the annotated lines, we evaluate our system by checking the error-rate obtained by using Levenshtein distance comparing each line from the automatically analyzed poem against a hand-made scansion from the Gold Stan-

---

[10] https://bitbucket.org/mhulden/pyperceptron
[11] https://github.com/jakevdp/pyCRFsuite
[12] https://github.com/glample/tagger
[13] See the machine_translation example at https://github.com/mila-udem/blocks-examples
[14] https://github.com/karpathy/char-rnn/
[15] In this description, the elements in a sequence can be either words or syllables, separated by spaces.

[16] Syllable to Stress.
[17] Word to Stress Pattern.
[18] We saw slight improvements in the results by including pre-trained word embeddings in the English dataset, but improvements were not significant.

|  | **English** | | **Spanish** | |
| --- | --- | --- | --- | --- |
|  | Per Syllable (%) | Per Line (%) | Per Syllable (%) | Per Line (%) |
| ZeuScansion (Agirrezabal et al., 2016b) | 86.78 | 26.21 | - | - |
| Scandroid (Hartman, 2005) | 89.78 | 42.95 | - | - |
| Gervas (2000)* | - | - | - | 88.73 |
| Perceptron$_{10}$ (S2S) | 84.86 | 29.32 | 74.54 | 0.31 |
| Perceptron$_{64}$ (S2S) | 89.34 | 43.36 | 92.25 | 40.78 |
| HMM (S2S) | 90.43 | 49.88 | 92.57 | 45.40 |
| CRF$_{10}$ (S2S) | 89.66 | 50.16 | 85.30 | 19.20 |
| CRF$_{64}$ (S2S) | 91.41 | 55.30 | 93.37 | 57.00 |
| Bi-LSTM+CRF (S2S) | 91.26 | 55.28 | 95.13 | 63.68 |
| Bi-LSTM+CRF+WB (S2S) | **92.96** | **61.39** | 98.74 | 88.82 |
| Bi-LSTM+CRF (W2SP) | 89.39 | 44.29 | **98.95** | **90.84** |

Table 1: Results of the classifiers in the English and Spanish datasets. The first lines show three rule-based scansion systems, ordered according to their overall performance. The next group displays the results presented in Agirrezabal et al. (2016a) (gray cells) together with the results on the Spanish dataset under the same conditions. The last three lines show the accuracies of the neural network architecture.

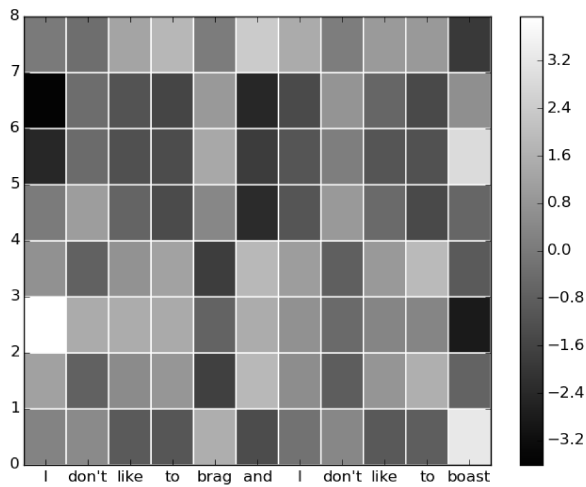

Figure 1: Output layer activations of the Bidirectional LSTM (8 different outputs), which will be the input of the CRF layer. The input sentence is "*I don't like to brag and I don't like to boast*".

In figure 1 the output of the Bidirectional LSTMs can be seen, showing the first line of "*Scrambled Eggs Super!*". The columns that represent the stressed syllables (2nd, 5th, 8th and 11th syllables) stand out clearly. This representation is used as input for the CRF layer, which finds the optimal resulting sequence.

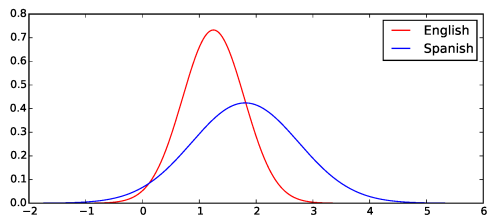

Figure 2: Syllable length distribution in Spanish and English.

## 6 Discussion and Future Work

Inspecting the results in table 1, we conclude that the Spanish dataset is more regular, as results generally are better. In the work by Agirrezabal et al. (2016a) 10 potentially language-agnostic features were used, and the extrapolation to Spanish seems to show that these features are not well-suited as language-agnostic features because of their small impact on the results. This claim should be better demonstrated by performing further experiments in more languages.

Also, it can be inferred that word structure in Spanish plays a key role, mainly because the inclusion of word boundaries (or just word informa-dard (Graves, 2012, p. 13). We do this in order to avoid overly penalizing a single missing or superfluous syllable, something which would shift an entire pattern to the left or right and potentially produce a count where all syllables would be counted as incorrect even though only one syllable was missing or added.

Table 1 shows the results obtained compared with the ones presented in Agirrezabal et al. (2016a), by applying the same methods with the same parameters. Results from that work are shown with a gray colored background.

tion) improves results significantly compared with systems that do not. In table 1, the Perceptron-based results are significantly improved upon by using 64 feature templates (according to a Welch's two-sample t-test ($p < 0.05$)).[19] The same can be seen in the case of CRFs and also when word boundary information is provided to the Bidirectional LSTM. The importance of knowing word boundaries in Spanish can be attributed to the fact that while English words tend to be monosyllabic, not all Spanish words will contain a syllable boundary marker, making it an informative characteristic. Figure 2 shows the differences of English and Spanish words' average syllable length. The results of the Bidirectional LSTM show that if tagged data is available, very good results can be obtained with neural network based structured predictors, without the use of additional linguistic information (such as, lexical stress, POS-tags, etc.). Tentatively, it could be said that the models infer the phonological information inherent to the words, although showing this conclusively requires further experiments.

The results serve to prompt several new strands of research in the domain. Our main goal is to be able to analyze poems with minimal supervision, including knowledge of the language in question, with unsupervised learning of rhythmic patterns being our long-term goal, possibly extending the unsupervised work done in Greene et al. (2010). We also intend to use the neural network based metrical analyzer as a meter checker in an automatic poetry generation system, such as Manurung (2003); Toivanen et al. (2013); Gervás (2014); Oliveira et al. (2014).

## Acknowledgments

The first author's work has been partially funded by the University of the Basque Country (UPV/EHU) in collaboration with the Association of the Friends of Bertsolaritza under the Zabalduz program. We also want to acknowledge the anonymous reviewers, as their feedback improved both the current paper and contained suggestions for future research.

## References

Manex Agirrezabal, Iñaki Alegria, and Mans Hulden. 2016a. Machine Learning for Metrical Analysis of English Poetry. In *Proceedings of COLING 2016, the 26th International Conference on Computational Linguistics, Osaka, Japan*. pages 772–781.

Manex Agirrezabal, Aitzol Astigarraga, Bertol Arrieta, and Mans Hulden. 2016b. ZeuScansion: a Tool for Scansion of English Poetry. *Journal of Language Modelling* 4(1):3–28.

Dzmitry Bahdanau, Kyunghyun Cho, and Yoshua Bengio. 2014. Neural Machine Translation by Jointly Learning to Align and Translate. *arXiv preprint arXiv:1409.0473* .

José Domínguez Caparrós. 1999. *Diccionario de Métrica Española*. Alianza Editorial.

Alfred Corn. 1997. *The Poem's Heartbeat: A Manual of Prosody.*. Copper Canyon Press.

Alex Estes and Christopher Hench. 2016. Supervised [Machine Learning for Hybrid Meter. *on Computational Linguistics for Literature* page 1.

Nigel Fabb. 1997. *Linguistics and Literature : Language in the Verbal Arts of the World*. Blackwell. 2nd ISBN: 9780631192435. http://strathprints.strath.ac.uk/28825/.

Yoav Freund and Robert E. Schapire. 1999. Large Margin Classification Using the Perceptron Algorithm. *Machine learning* 37(3):277–296.

Pablo Gervas. 2000. A Logic Programming Application for the Analysis of Spanish Verse. In *Computational Logic—CL 2000*, Springer, pages 1330–1344.

Pablo Gervás. 2014. Composing Narrative Discourse for Stories of many Characters: A Case Study over a Chess Game. *Literary and Linguistic Computing* page fqu040.

Alex Graves. 2012. *Supervised Sequence Labelling with Recurrent Neural Networks*. Springer.

Erica Greene, Tugba Bodrumlu, and Kevin Knight. 2010. Automatic Analysis of Rhythmic Poetry with Applications to Generation and Translation. In *Proceedings of the 2010 Conference on Empirical Methods in Natural Language Processing*. Association for Computational Linguistics, pages 524–533.

Péter Halácsy, András Kornai, and Csaba Oravecz. 2007. HunPos: an open source trigram tagger. In *Proceedings of the 45th Annual Meeting of the ACL on Interactive Poster and Demonstration Sessions*. Association for Computational Linguistics, pages 209–212.

Charles O. Hartman. 2005. The Scandroid 1.1. http://oak.conncoll.edu/cohar/Programs.htm.

Bruce Hayes, Colin Wilson, and Anne Shisko. 2012. Maxent Grammars for the Metrics of Shakespeare and Milton. *Language* 88(4):691–731.

---

[19] The current word is one of the feature templates.

Malcolm Hayward. 1996. Analysis of a Corpus of Poetry by a Connectionist Model of Poetic Meter. *Poetics* 24(1):1–11.

Katharina Kann and Hinrich Schütze. 2016. MED: The LMU system for the SIGMORPHON 2016 Shared Task on Morphological Reinflection. *ACL 2016* page 62.

John Lafferty, Andrew McCallum, and Fernando C. N.H Pereira. 2001. Conditional Random Fields: Probabilistic Models for Segmenting and Labeling Sequence Data .

Guillaume Lample, Miguel Ballesteros, Sandeep Subramanian, Kazuya Kawakami, and Chris Dyer. 2016. Neural Architectures for Named Entity Recognition. In *Proceedings of NAACL-2016, San Diego, California, USA*. Association for Computational Linguistics.

Harry M Logan. 1988. Computer Analysis of Sound and Meter in Poetry. *College Literature* pages 19–24.

Ruli Manurung. 2003. *An evolutionary algorithm approach to poetry generation*. Ph.D. thesis, School of informatics, University of Edinburgh.

G McAleese. 2007. *Improving Scansion with Syntax: an Investigation into the Effectiveness of a Syntactic Analysis of Poetry by Computer using Phonological Scansion Theory*. Ph.D. thesis, Open University.

Tomas Mikolov, Martin Karafiát, Lukas Burget, Jan Cernockỳ, and Sanjeev Khudanpur. 2010. Recurrent Neural Network based Language Model. In *Interspeech*. volume 2, page 3.

Borja Navarro-Colorado. 2015. A Computational Linguistic Approach to Spanish Golden Age Sonnets: Metrical and Semantic Aspects. *Computational Linguistics for Literature* page 105.

Borja Navarro-Colorado, Marıa Ribes Lafoz, and Noelia Sánchez. 2016. Metrical Annotation of a Large Corpus of Spanish Sonnets: Representation, Scansion and Evaluation. In *Proceedings of the Language Resources and Evaluation Conference*.

Naoaki Okazaki. 2007. CRFsuite: a fast implementation of Conditional Random Fields (CRFs). http://www.chokkan.org/software/crfsuite/.

Hugo Gonçalo Oliveira, Raquel Hervás, Alberto Díaz, and Pablo Gervás. 2014. Adapting a Generic Platform for Poetry Generation to Produce Spanish Poems. *International Conference on Computational Creativity* .

Marc R Plamondon. 2006. Virtual Verse Analysis: Analysing Patterns in Poetry. *Literary and Linguistic Computing* 21(suppl 1):127–141.

Antonio Quilis. 1984. *Métrica Española*. Ariel Barcelona.

Lawrence R. Rabiner. 1989. A Tutorial on Hidden Markov Models and selected Applications in Speech Recognition. *Proceedings of the IEEE* 77(2):257–286.

Frank Rosenblatt. 1958. The Perceptron: a Probabilistic Model for Information Storage and Organization in the Brain. *Psychological review* 65(6):386.

Timothy Steele. 1999. *All the Fun's in how you Say a Thing: an Explanation of Meter and Versification*. Ohio University Press Athens.

Jukka Toivanen, Matti Järvisalo, and Hannu Toivonen. 2013. Harnessing Constraint Programming for Poetry. *International Conference on Computational Creativity* .

Navarro Tomás Tomás. 1995. *Métrica Española*. Ed. Labor.

Herbert F Tucker. 2011. Poetic Data and the News from Poems: A for Better for Verse Memoir. *Victorian Poetry* 49(2):267–281.